\documentclass[journal]{IEEEtran}

\usepackage{amsmath}
\usepackage{amssymb}
\usepackage{booktabs}
\usepackage{cite}
\usepackage{graphicx}
\usepackage{hyperref}
\usepackage{multirow}

\newcommand{\mrr}{\operatorname{MRR}}
\newcommand{\logit}{\operatorname{logit}}

\begin{document}

\title{Back to All-Entity Ranking: Sampler-Dependent Evaluation in
Continuous-Time Dynamic Graphs}

\author{Minwoo Yu and Young-guk Ha
\thanks{The authors are with the Smart Computing Laboratory, Department of
Computer Science and Engineering, Konkuk University, Seoul 05029, Republic of
Korea (e-mail: snowypainter@konkuk.ac.kr; ygha@konkuk.ac.kr).}
\thanks{Young-guk Ha is the corresponding author.}}

\maketitle

\begin{abstract}
Next-destination prediction in continuous-time dynamic graphs (CTDGs)
commonly ranks an observed interaction against sampled negative
destinations. The resulting score is conditional on both the negative
distribution and the number of candidates chosen by the researcher. We show
that a non-uniform negative distribution changes the Bayes-optimal ranking,
while even a finite candidate set drawn uniformly can destabilize model
rankings and measured module effects.

Time-varying source--destination history membership and model operations that
use this information directly transmit the sampler's influence to the
evaluation score. We examine this mechanism using a factorial evaluation of
repeated and new positives against seen and unseen negatives, a minimal
scorer based solely on pair-history membership, and controlled representation
interventions. Across six models on LastFM, MOOC, Reddit, and Wikipedia, at
least one model pair changes relative order between the expected Uniform-20
metric and the full catalog on three of the four datasets. The measured
effect of the same module also changes in magnitude and direction with the
candidate-set size and training objective.

These results establish that model-superiority and ablation conclusions from
sampled-negative benchmarks are conditional on the stated candidate
configuration. All-entity ranking evaluates every destination in a fixed
catalog, eliminating negative-selection freedom and sampling variation while
retaining the original CTDG scorer. We therefore recommend all-entity ranking
as the primary evidence for architecture comparisons on CTDG benchmarks with
an enumerable, fixed destination catalog.
\end{abstract}


\section{Introduction}
\label{sec:introduction}

A continuous-time dynamic graph (CTDG) observes an interaction stream
\(e_i=(u_i,v_i,t_i)\) in chronological order and predicts the next
destination from the history available at the current time. A common
evaluation ranks the observed destination \(v^+\) together with sampled
negative destinations
\cite{rossi2020tgn,poursafaei2022dgb,huang2023tgb}. The evaluator draws
\(K\) negative destinations from
\begin{equation}
v_1^-,\ldots,v_K^-\sim q(v\mid u,t,\mathcal H_t),
\label{eq:qk_intro}
\end{equation}
where \(q\) is the probability distribution that selects a negative
destination given source \(u\), time \(t\), and past history
\(\mathcal H_t\), and \(K\) is the number of negatives compared with one
positive. Random-20, for example, uses a \(q\) that draws destinations
uniformly from a fixed candidate space and sets \(K=20\).

This study considers CTDG next-destination prediction in which source and
destination roles are distinct and the destination catalog defined by the
training prefix is shared at test time. The JODIE interaction datasets
Wikipedia, Reddit, MOOC, and LastFM follow this setting. Open-world graphs
with rapidly growing node sets, homogeneous graphs in which source and
destination roles are interchangeable, and tasks with event-specific
eligibility require separate risk-set definitions.

The Dynamic Graph Benchmark (DGB) established that uniform random negatives
contain many easy edges and that historical and inductive negatives alter
both model performance and model order \cite{poursafaei2022dgb}. The Temporal
Graph Benchmark (TGB) and subsequent evaluation studies likewise reported
changes in difficulty and ranking under different candidate constructions
\cite{huang2023tgb,romero2023perspectives,gastinger2024tgb2}. In
recommendation, sampled metrics have been shown theoretically and
empirically to disagree with full-item rankings and potentially fail to
preserve the relative order of models
\cite{krichene2020sampled,dallmann2021sampling}.

We connect this established evaluation discrepancy to time-dependent pair
history and architecture conclusions in CTDGs. Deriving the Bayes-optimal
score for sampled ranking shows that a non-uniform \(q\), whose probability
varies across candidates, orders destinations by
\(p_+(v\mid x)/q(v\mid x)\). A uniform \(q\) over a fixed catalog preserves
the Bayes-optimal full-catalog order, but MRR measured with finite \(K\)
observes only a nonlinear partial view of that order. Consequently, the same
checkpoint and scorer can yield different MRR values, model rankings, and
module effects as \(q\) and \(K\) change.

In CTDGs, this phenomenon interacts with time-varying pair membership. Pair
membership is the binary indicator of whether the current source and a
candidate destination have interacted previously. Recent models jointly
represent the histories of two nodes or directly inject recurrence history
for the current pair into the scoring path
\cite{yu2023dygformer,tian2024tncn,yi2025craft,yu2025tami}. In a small random
candidate set, a positive is often a previously observed pair whereas the
negatives are previously unseen pairs. Coarse separation of the seen and
unseen groups can then produce a high rank, while the ability to choose the
actual destination within either group remains conflated in a single MRR.

We examine this connection in four steps. First, we derive the Bayes-optimal
ranking from the posterior over a sampled candidate set and distinguish
uniform from non-uniform \(q\). Second, we compare the optimal coefficient of
a minimal scorer based only on pair-history membership with the coefficient
learned from data. Third, a fully crossed \(2\times2\) evaluation of
repeated/new positives and seen/unseen negatives, together with an
orthogonal representation projection, separates subgroup separation from
within-group ranking. Fourth, using identical score vectors from six CTDG
models, we compute sampled MRR for
\(K\in\{1,5,20,50,100\}\) and MRR over the entire fixed training catalog.

The principal empirical result is that model ranking and measured module
effect depend on \(q\) and \(K\). On LastFM, expected Uniform-20 ranks CRAFT
above DyGFormer, whereas the full catalog ranks DyGFormer above CRAFT. The
ordering of the three leading models also changes on MOOC. On Wikipedia, the
expected gain of CRAFT-R over CRAFT moves from \(+0.0037\) with 20 negatives
to \(-0.0266\) over the full catalog. The former has a seed-paired confidence
interval containing zero, while the latter is negative. Beyond the
arithmetic decline in MRR caused by enlarging a candidate set, the conclusion
about which model or design component is superior can therefore change.

This finding also bears directly on causal explanations of
sampled-negative improvements. CRAFT attributes the advantage of
candidate-to-history cross-attention to learning compatibility between the
destination and the source history \cite{yi2025craft}. Yet the same operation
that queries history with candidate identity also conveys whether the
source--destination pair has appeared before. Under random candidates, these
two signals often point toward the correct answer in the same direction, so
sampled MRR alone cannot identify their respective contributions. Our
membership controls and representation interventions isolate this
alternative path, while all-entity ranking enables architecture comparison
without relying on a sampled candidate composition.

\begin{itemize}
\item \textbf{Theory.} A non-uniform \(q\) changes the Bayes target, whereas
uniform finite-\(K\) evaluation destabilizes model and module comparisons.
\item \textbf{Mechanism.} Time-varying pair membership and pair-history
operations amplify this effect.
\item \textbf{Recommendation.} Architecture comparisons on CTDG
next-destination benchmarks with an enumerable fixed catalog should use
all-entity ranking as their primary evidence.
\end{itemize}

\section{Related Work}
\label{sec:related_work}

\subsection{CTDG Link Prediction and Negative Sampling}

TGN applies binary cross-entropy to observed edges and sampled negative edges
\cite{rossi2020tgn}. DGB organized random, historical, and inductive
negatives within a unified dynamic link-prediction setting and reported that
the simple EdgeBank baseline performs strongly against easy random negatives
\cite{poursafaei2022dgb}. DyGLib applied these three distributions to
multiple CTDG architectures \cite{yu2023dygformer}, while TGB provided an
MRR-based evaluation comparing one positive with multiple negatives
\cite{huang2023tgb}. TGB 2.0 compared random 1-vs-1000,
destination-aware 1-vs-1000, and 1-vs-all evaluation, demonstrating the
effect of the candidate space on reported results \cite{gastinger2024tgb2}.

The Stanford JODIE datasets Wikipedia, Reddit, MOOC, and LastFM remain widely
used in CTDG research. The original JODIE evaluation compared the next item
with every item in the dataset and reported MRR and Recall@10 for Reddit,
Wikipedia, and LastFM \cite{kumar2019jodie}. These interactions were
subsequently reformulated widely as sampled-negative binary link prediction.
RepeatMixer and TPNet evaluated recurrence information and pairwise temporal
features on the four datasets using random, historical, and inductive
AP/AUC \cite{zou2024repeatmixer,lu2024tpnet}. Yang et al.\ compared samplers,
aggregators, and memory modules for temporal graph neural networks at scale
using \(1{:}1\) training, \(1{:}9\) validation, and \(1{:}49\) test
candidates \cite{yang2025ideal}. BandRank optimized listwise MRR and Hits@5
over query-specific candidate sets \cite{li2025bandrank}. These studies
refine the units used to compare CTDG architectures, and each conclusion is
conditional on its stated candidate construction.

\subsection{Sampled and Full-Catalog Metrics}

Recommendation research has analyzed the discrepancy between sampled and
full-catalog metrics directly. Krichene and Rendle showed that a sampled
metric may fail to preserve the relative ordering of models under the exact
metric even in expectation \cite{krichene2020sampled}. Dallmann et al.\
reported that model orders obtained for sequential recommenders under
uniform and popularity sampling do not consistently agree with full ranking
\cite{dallmann2021sampling}. We connect this general result to
time-dependent pair membership, pair-history operations, and
sampler-specific optimal reward directions in CTDGs.

All-entity evaluation also has a clear lineage in CTDGs. DyRep computed a
rank by replacing the test destination with every other entity
\cite{trivedi2019dyrep}, and JODIE compared the ground-truth item with every
item \cite{kumar2019jodie}. Following this lineage, we apply contemporary
CTDG scorers to every destination in a fixed candidate space. The original
scorer is retained, while the freedom to choose \(q\) and \(K\) is removed
from architecture and module comparisons.

\subsection{CTDG Models Using Pair History}

TGN updates node-specific memory and combines the representations of two
nodes in the final link predictor \cite{rossi2020tgn}. GraphMixer similarly
encodes each node's temporal neighbors independently before predicting a
link from the two representations \cite{cong2023graphmixer}. DyGFormer
computes neighbor co-occurrence between the historical sequences of the
source and destination \cite{yu2023dygformer}. TNCN and NAT use temporal
common neighbors or joint neighborhood relations in a pair representation
\cite{tian2024tncn,luo2022nat}.

CRAFT uses the candidate destination as a query to compress the source's
recent history through cross-attention. CRAFT-R augments this representation
with the number of previous occurrences of the current source--destination
pair \cite{yi2025craft}. Link History Aggregation (LHA) in TAMI is an adapter
that separately retains recent interactions of the current pair; in our
experiments, it is attached to DyGFormer \cite{yu2025tami}. These models
provide expressive representations of pair compatibility and recurrence.
They also create direct paths through which sampler-induced differences in
seen/unseen composition can reach the score.

\subsection{Mechanistic Attribution of the Membership-Aligned Component}

CRAFT presents learnable node embeddings and target-aware matching as two
key requirements for future-link prediction. The former provides persistent
node identities, while the latter directly computes compatibility between a
destination and the source behavior history through candidate-to-history
cross-attention. CRAFT-R additionally incorporates pair recurrence counts in
settings with many seen edges. CRAFT supports this design through
sampled-negative performance and comparisons with existing backbones
augmented by learnable embeddings \cite{yi2025craft}.

When candidate identity forms the query and past destination identities form
the keys, cross-attention can encode both semantic compatibility and an
identity match with a historical pair. If a random candidate set frequently
contains seen positive pairs and unseen negative pairs, the two signals
induce the same ranking direction. Sampled MRR then observes their combined
effect. Our \(2\times2\) membership control and representation intervention
trace the membership-aligned component of CRAFT's predictive performance.
Residual ranking performance after removing membership-aligned separation
captures the remaining predictive signals jointly.

\section{The Ranking Target of Sampled-Negative Evaluation}
\label{sec:theory}

\subsection{Problem Setup}

Let \(x=(u,t,\mathcal H_t)\) denote the context at time \(t\),
\(p_+(v\mid x)\) the conditional distribution of the observed destination,
and \(q(v\mid x)\) the negative distribution. Construct
\(C=\{v_0,\ldots,v_K\}\) from one positive and \(K\) independent negatives,
with an equal prior probability that each position contains the positive.
The posterior probability that candidate \(j\) is positive is
\begin{align}
\Pr(J=j\mid C,x)
&=
\frac{p_+(v_j\mid x)\prod_{i\ne j}q(v_i\mid x)}
{\sum_{\ell=0}^{K}p_+(v_\ell\mid x)
\prod_{i\ne\ell}q(v_i\mid x)} \notag\\
&=
\frac{p_+(v_j\mid x)/q(v_j\mid x)}
{\sum_{\ell=0}^{K}p_+(v_\ell\mid x)/q(v_\ell\mid x)}.
\label{eq:posterior}
\end{align}
Equation~\eqref{eq:posterior} shows that this posterior is the normalized
value of \(p_+(v\mid x)/q(v\mid x)\). The Bayes-optimal ordering maximizes
expected reciprocal rank when the candidate-generating distribution is
known exactly. Placing candidates in descending posterior order achieves
this maximum. A standard log-score inducing the same order is
\begin{equation}
\boxed{\ell_q^*(x,v)=\log p_+(v\mid x)-\log q(v\mid x)}.
\label{eq:bayes_score}
\end{equation}
Thus, the optimal score for sampled ranking is the log-density ratio between
the observed-destination and negative distributions. This result agrees with
noise-contrastive estimation, in which the optimal classifier separating
data from noise learns a density ratio \cite{gutmann2010nce}. Proper-scoring
rule theory formalizes the property that a log score elicits the stated
probability distribution truthfully \cite{gneiting2007proper}.

For sampling without replacement, let \(Q\) be the joint probability of the
negative set. The posterior becomes
\begin{equation}
\Pr(J=j\mid C,x)\propto
p_+(v_j\mid x)Q(C\setminus\{v_j\}\mid x,v_j).
\label{eq:without_replacement}
\end{equation}
In~\eqref{eq:without_replacement}, the product \(\prod q\) for independent
sampling is replaced by the actual sampler \(Q\); the candidate-generation
rule remains part of the Bayes-optimal ordering.

\subsection{Uniform Finite-\texorpdfstring{\(K\)}{K} and Non-Uniform Samplers}

For a fixed catalog \(\mathcal R\), the uniform sampler is
\(q_{\mathrm{unif}}(v\mid x)=1/|\mathcal R|\). Its optimal score is
\begin{equation}
\ell_{\mathrm{unif}}^*(x,v)
=\log p_+(v\mid x)+\log|\mathcal R|.
\label{eq:uniform_score}
\end{equation}
The second term in~\eqref{eq:uniform_score} is constant across candidates.
The population-optimal ordering under uniform sampling therefore equals the
optimal full-catalog ordering. Sampled MRR, however, observes only a random
subset of the full order. Averaging a nonlinear rank metric across scorers
can make not only its absolute value but also relative model order and module
gain depend on \(K\) \cite{krichene2020sampled,dallmann2021sampling}.

Under uniform sampling without replacement, the expected metric can be
computed exactly without a particular negative draw. Let \(r\) be the
positive's full-catalog rank and \(N\) the catalog size. The number \(X\) of
the \(K\) negatives scoring above the positive follows
\begin{equation}
X\sim\operatorname{Hypergeom}(N-1,r-1,K),
\qquad \operatorname{RR}_K=\frac{1}{1+X}.
\label{eq:hypergeom_rank}
\end{equation}
Its expected reciprocal rank is
\begin{equation}
\mathbb E[\operatorname{RR}_K\mid r,N]
=
\frac{
\binom{N}{K+1}-\binom{N-r}{K+1}
}{
r\binom{N-1}{K}
}.
\label{eq:expected_sampled_rr}
\end{equation}
Equation~\eqref{eq:expected_sampled_rr} determines expected finite-\(K\) MRR
solely from each event's full rank. It removes evaluation-seed variance and
separates changes caused by a particular candidate draw from those caused by
the finite-\(K\) metric transformation. At \(K=N-1\), the expectation becomes
\(1/r\), equal to the all-entity reciprocal rank.

For historical, source-history, or popularity samplers,
\(q(v\mid x)\) varies by candidate, so the term \(-\log q(v\mid x)\)
in~\eqref{eq:bayes_score} directly affects the order. The support of a
source-history sampler is
\begin{equation}
\operatorname{supp}q_{\mathrm{hist}}(\cdot\mid x)
=\mathcal H_u(t)\setminus\{v^+\}.
\label{eq:history_support}
\end{equation}
For a new positive in~\eqref{eq:history_support}, the positive lies outside
the support while every negative lies inside it. Pair-history membership
then separates the labels perfectly. A non-uniform sampler consequently
determines not only candidate difficulty but also the optimal ordering
rewarded by evaluation.

\subsection{Optimal Coefficient of Pair Membership}

Define pair membership as
\begin{equation}
M(u,v,t)=\mathbb 1[(u,v)\in\mathcal H_t].
\label{eq:membership}
\end{equation}
Equation~\eqref{eq:membership} indicates whether source \(u\) and destination
\(v\) have interacted at least once before time \(t\). Consider the minimal
scorer \(s_\beta(M)=b+\beta M\), which uses only this indicator. Let \(p_1\)
and \(q_1\) be the probabilities that \(M=1\) among positives and negatives,
respectively. The optimal seen/unseen score difference is
\begin{equation}
\boxed{\beta_q^*=\logit(p_1)-\logit(q_1)}.
\label{eq:membership_beta}
\end{equation}
Equation~\eqref{eq:membership_beta} sets the optimal reward to the difference
in log odds between the seen rate of true positives and that induced among
negatives by the sampler.

Under uniform random sampling,
\(q_1=|\mathcal R_{\mathrm{seen}}(u,t)|/|\mathcal R|\); if \(p_1>q_1\), then
\(\beta_q^*>0\). This reflects useful recurrence in the data: positive events
are more concentrated on historical pairs than their catalog prevalence
would suggest. Under source-history sampling, \(q_1=1\), and in the presence
of new positives, \(\beta_q^*\rightarrow-\infty\). This limit expresses the
opposite perfect separation created by the sampler support. The concern is
therefore not recurrence itself. Rather, coarse seen/unseen separation can
account for a large share of MRR in a small candidate set and obscure the
ability to select a relation within the seen or unseen subgroup.

\section{Controlled Evaluation and the All-Entity Protocol}
\label{sec:protocol}

\subsection{\texorpdfstring{\(2\times2\)}{2x2} Pair-Membership Evaluation}

Fully crossing the historical membership of positive and negative pairs
produces the four conditions in Table~\ref{tab:2x2_design}. Repeated--Seen is
recurrent-link ranking: selecting the relation that recurs now among
relations observed previously. New--Unseen is novel-link ranking: selecting
the actual positive among relations being formed with the source for the
first time. In both conditions, positive and negative membership agree, so
membership alone cannot separate the labels.

\begin{table*}[t]
\centering
\caption{Evaluation design crossing positive and negative pair membership.}
\label{tab:2x2_design}
\begin{tabular}{lll}
\toprule
 & Seen negative & Unseen negative\\
\midrule
Repeated positive & Recurrent-link ranking & Seen-subgroup separation\\
New positive & Unseen-subgroup separation & Novel-link ranking\\
\bottomrule
\end{tabular}
\end{table*}

In Repeated--Unseen, the positive pair has appeared before; in New--Seen, the
positive pair appears for the first time. These two conditions diagnose
whether a model assigns higher scores to the seen or unseen subgroup. An
unseen negative forms a new pair with the current source, but its destination
node must already have appeared before \(t\). We apply the same checkpoint to
the same positive events for which all four conditions can be constructed,
changing only the negative composition. The two off-diagonal cells use 20
negatives to diagnose membership direction. For the primary diagonal
evaluation, a repeated positive is compared with every seen destination of
its source, while a new positive is compared with every unseen destination
in the fixed catalog. This all-seen/all-unseen decomposition removes
finite-\(K\) sampling from the diagonal cells as well.

\subsection{Pair-Membership Representation Intervention}

The pre-scorer representation of CRAFT-R, which uses pair history explicitly,
is
\begin{equation}
z_{uv}(t)=[h_{u\mid v}(t);e_v(t);r_{uv}(t)],
\label{eq:craft_representation}
\end{equation}
combining a candidate-conditioned source representation, a candidate
embedding, and a pair-recurrence representation. Define the difference
between the mean representations of seen and unseen training pairs as
\begin{equation}
a=\mu_{\mathrm{seen}}-\mu_{\mathrm{unseen}}.
\label{eq:membership_direction}
\end{equation}
We remove this direction using the orthogonal projection
\begin{equation}
P_\perp=I-\frac{aa^\top}{a^\top a},
\qquad z^\perp_{uv}(t)=P_\perp z_{uv}(t).
\label{eq:projection}
\end{equation}
Equation~\eqref{eq:projection} removes only the component parallel to \(a\)
and preserves its orthogonal complement. This is a controlled intervention
for tracing the representation path behind an observed score change, rather
than a new prediction method. As controls, we also remove a random direction
of equal dimensionality and a direction computed after shuffling the
membership labels, thereby measuring the effect of removing an arbitrary
single dimension.

\subsection{Fixed-Catalog All-Entity Ranking}

The primary risk set \(\mathcal R\) is fixed to all destination identities
that occur at least once in the chronological training prefix. We exclude
the source itself and, for test events whose positive belongs to
\(\mathcal R\), evaluate the model's existing score for every destination.
Full-catalog MRR averages the reciprocal rank of the observed destination
within the entire catalog across test events. Hits@10 is computed from the
same complete order. We assign average rank to score ties.

All-entity ranking requires neither a separate intensity head nor probability
calibration. Any CTDG model returning a source--destination score can be
evaluated using the same checkpoint and scorer. One complete score vector per
event yields both the full-catalog and sampled results. Sampled MRR uses the
first \(K\in\{1,5,20,50,100\}\) elements of a uniform random permutation
excluding the positive. This nested construction shares candidates across
different values of \(K\). The primary \(K\)-sweep applies
Eq.~\eqref{eq:expected_sampled_rr} to each event's full rank to compute the
expected MRR under uniform sampling. Actual nested draws are retained as a
secondary check against the expectation curve.

For a stateful model, we score every candidate from the same current memory
and update memory exactly once with the observed positive interaction. This
procedure prevents leakage caused by candidate-specific state changes. A
secondary sensitivity analysis forms a time-available catalog from
destinations that have appeared at least once before each event.

\subsection{Comparison Quantities}

We measure agreement between the sampled and full-catalog model orders using
Kendall's \(\tau\). A value of \(\tau=1\) means that every model pair has the
same relative order, while \(\tau=-1\) means that every order is reversed.
We measure the effect of the pair-recurrence module in CRAFT-R as
\begin{equation}
G_R(K)=\mrr_K(\text{CRAFT-R})-\mrr_K(\text{CRAFT}).
\label{eq:module_gain}
\end{equation}
The magnitude and sign of \(G_R(K)\) report the change recorded by evaluation
when the module is added to the same backbone.

\section{Experiments}
\label{sec:experiments}

\subsection{Setup}

Table~\ref{tab:experiment_settings} summarizes the common training and
evaluation settings.
\begin{table*}[t]
\centering
\caption{Training and evaluation settings.}
\label{tab:experiment_settings}
\small
\begin{tabular}{ll}
\toprule
Item & Setting\\
\midrule
Datasets & LastFM, MOOC, Reddit, Wikipedia\\
Event range & Most recent 32,768 events of each dataset\\
Temporal split & Training prefix 85\%, test suffix 15\%\\
History length & 320\\
Training epochs & 3\\
Seeds & 7, 17, 27\\
Primary training objective & BCE, one uniform random negative per positive\\
Objective sensitivity & CRAFT/CRAFT-R retrained with BPR\\
Models & CRAFT, CRAFT-R, DyGFormer, DyGFormer+LHA, TGN, GraphMixer\\
Primary catalog & All destinations appearing in the training prefix\\
Sampled evaluation & \(K\in\{1,5,20,50,100\}\), nested uniform candidates\\
Catalog coverage & LastFM 1.000, MOOC 1.000, Reddit 0.996, Wikipedia 0.987\\
Reported values & Mean and standard deviation over three seeds\\
\bottomrule
\end{tabular}
\end{table*}

We compare CRAFT, CRAFT-R, DyGFormer, DyGFormer+LHA, TGN, and GraphMixer.
TGN and GraphMixer encode each node history separately and combine the two
nodes only in the final scorer. DyGFormer uses co-occurrence between the two
histories. In CRAFT, the candidate enters as the cross-attention query over
source history, and CRAFT-R adds pair recurrence. The LHA condition attaches
TAMI's pair-history adapter to DyGFormer. All models are trained with BCE
using one random negative; parameters are frozen during evaluation. We
report the mean and standard deviation over three seeds.

This comparison is a unified-objective architecture study that matches
training conditions across architectures. We additionally train CRAFT and
CRAFT-R with the BPR objective used in the original CRAFT study, separating
training-objective sensitivity of module gain from stability of the
membership mechanism.

\subsection{Reward Direction Predicted by the Minimal Scorer}

Table~\ref{tab:beta} tests Eq.~\eqref{eq:membership_beta} under paired-random
and source-history conditions on the same events. The analytic \(\beta^*\)
and directly learned \(\beta\) agree on every dataset under random sampling.
Under source-history sampling, the coefficient moves to a large negative
value. The positive random coefficient reflects recurrence in the data,
where positive events concentrate on seen pairs. The negative
source-history coefficient reflects opposite perfect separation created by
the sampler support.

\begin{table*}[t]
\centering
\caption{Analytic and learned values for the minimal pair-membership scorer.}
\label{tab:beta}
\scriptsize
\begin{tabular}{lrrrrrr}
\toprule
Dataset & Events & \(p_1\) & Random \(q_1\) & \(\beta^*\) & Random \(\beta\) & History \(\beta\)\\
\midrule
LastFM & 18,366 & .816 & .122 & 3.464 & 3.464 & -8.606\\
Wikipedia & 971 & .719 & .093 & 3.211 & 3.211 & -8.732\\
Reddit & 880 & .828 & .062 & 4.290 & 4.290 & -8.590\\
MOOC & 10,195 & .607 & .344 & 1.078 & 1.078 & -8.902\\
\bottomrule
\end{tabular}
\end{table*}

\subsection{\texorpdfstring{\(2\times2\)}{2x2} Results}

Models that directly transmit pair history show large amplitudes in the two
subgroup-separation conditions. CRAFT-R obtains Repeated--Unseen MRR of
0.957, 0.979, 0.987, and 0.888 on LastFM, Wikipedia, Reddit, and MOOC,
respectively. When only the positive is unseen (New--Seen), the corresponding
values are 0.059, 0.050, 0.048, and 0.508. DyGFormer+LHA likewise records
Repeated--Unseen MRR of 0.974--0.996 and New--Seen MRR of 0.048--0.050 on the
first three datasets. TGN and GraphMixer respond in the same direction, but
with smaller amplitudes.

The membership-matched diagonal provides separate relation-selection
results. CRAFT-R's recurrent-link MRR is 0.404, 0.431, 0.303, and 0.667 on
LastFM, Wikipedia, Reddit, and MOOC, while its novel-link MRR is 0.345,
0.303, 0.493, and 0.782. Subgroup separation and within-subgroup ranking,
which a single sampled MRR conflates, can therefore be read separately.

Removing the membership direction in Eq.~\eqref{eq:projection} reduces
Repeated--Unseen from 0.957 to 0.246 on LastFM, 0.979 to 0.201 on Wikipedia,
0.987 to 0.182 on Reddit, and 0.888 to 0.489 on MOOC. New--Seen moves from
0.059 to 0.657, 0.050 to 0.539, 0.048 to 0.772, and 0.508 to 0.940,
respectively. The changes in novel-link MRR are
\(-0.011,-0.070,-0.016,+0.004\). The intervention primarily changes the
preference between seen and unseen subgroups rather than selection among new
relations.

\begin{figure*}[t]
\centering
\includegraphics[width=\textwidth]{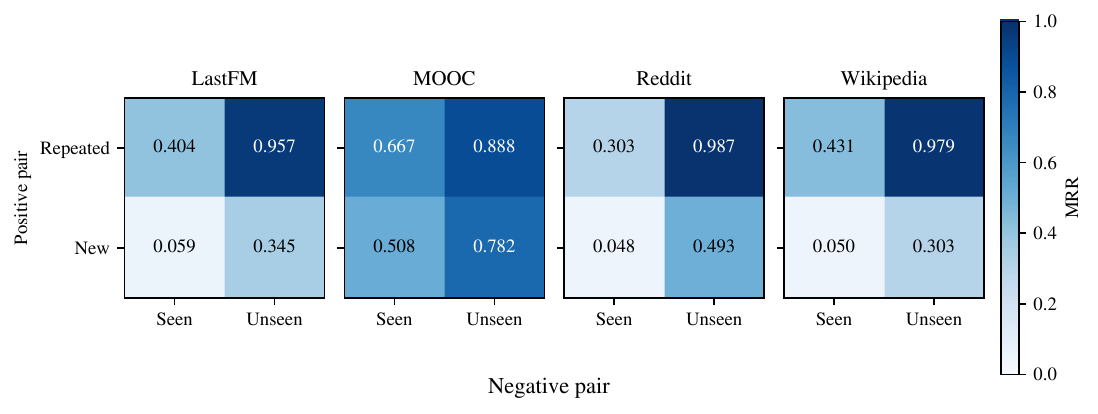}
\caption{Pair-membership \(2\times2\) MRR for CRAFT-R. Rows identify the
positive pair and columns identify the negative pair by historical
membership. All four cells are directional diagnostics using 20 negatives.}
\label{fig:craft_r_membership_2x2}
\end{figure*}

Figure~\ref{fig:craft_r_membership_2x2} displays the four CRAFT-R conditions
as a matrix. Rows distinguish repeated from new positive pairs, columns
distinguish seen from unseen negative pairs, and each cell reports MRR.
Every cell uses 20 negatives. The off-diagonal cells reverse the direction
in which pair-history membership identifies the answer and serve as
diagnostics.

The primary diagonal evaluation uses the exhaustive rankings in
Table~\ref{tab:membership_controlled_all_entity}. All-seen compares a repeated
positive with every destination previously encountered by its source.
All-unseen compares a new positive with every unseen destination in the fixed
catalog. CRAFT-R's all-unseen MRR is .0145 on LastFM, .5648 on MOOC, .1063 on
Reddit, and .0453 on Wikipedia.

\begin{table*}[t]
\centering
\caption{Membership-controlled all-seen and all-unseen MRR.}
\label{tab:membership_controlled_all_entity}
\scriptsize
\begin{tabular}{lrrrr}
\toprule
& \multicolumn{2}{c}{CRAFT} & \multicolumn{2}{c}{CRAFT-R}\\
\cmidrule(lr){2-3}\cmidrule(lr){4-5}
Dataset & All-seen & All-unseen & All-seen & All-unseen\\
\midrule
LastFM & .1939 & .0193 & .2041 & .0145\\
MOOC & .6253 & .5780 & .6247 & .5648\\
Reddit & .8715 & .0967 & .8549 & .1063\\
Wikipedia & .8923 & .0534 & .8202 & .0453\\
\bottomrule
\end{tabular}
\end{table*}

\subsection{Mechanistic Attribution of the Membership-Aligned Component}

CRAFT explains candidate-to-history cross-attention through fine-grained
compatibility between a destination and source behavior history
\cite{yi2025craft}. That explanation and pair-membership lookup predict the
same sampled-negative outcome: learnable candidate identity can match keys
in source history, and CRAFT-R passes the pair recurrence count directly to
the scorer.

The \(2\times2\) results distinguish these explanations. Holding CRAFT-R
parameters and event history fixed while changing only membership
composition yields Repeated--Unseen MRR of 0.888--0.987 and New--Seen MRR of
0.048--0.508. Orthogonally projecting out the membership direction strongly
reverses both diagnostics, whereas New--Unseen novel-link MRR changes by only
\(-0.070\) to \(+0.004\), depending on the dataset. The component localized
by the intervention therefore primarily separates the seen and unseen
subgroups rather than expressing compatibility among unseen candidates.

This result reproduces CRAFT's strong sampled MRR while giving its origin a
narrower interpretation. Cross-attention is a powerful
candidate-conditioned scorer, and sampled MRR jointly rewards semantic
matching and identity-based pair membership. Membership-matched
Repeated--Seen and New--Unseen rankings, together with all-entity ranking,
measure residual ranking performance after membership-aligned separation is
removed.

\subsection{Expected Uniform-20 and Full-Catalog Model Comparison}

Table~\ref{tab:main_result} reports MRR under expected Uniform-20 and the full
catalog using the same checkpoint, scorer, and test event. Reciprocal rank
declining as the candidate set grows is arithmetically expected; we therefore
distinguish this absolute decline from changes in relative model order and
module gain.

\begin{table*}[t]
\centering
\caption{Expected Uniform-20 MRR and MRR over the complete fixed training
catalog (mean \(\pm\) standard deviation).}
\label{tab:main_result}
\scriptsize
\begin{tabular}{llrrllrr}
\toprule
Dataset & Model & Expected-20 & Full & Dataset & Model & Expected-20 & Full\\
\midrule
LastFM & CRAFT & .4283$\pm$.0112 & .0636$\pm$.0065 &
MOOC & CRAFT & .7627$\pm$.0046 & .4625$\pm$.0066\\
& CRAFT-R & .6892$\pm$.0002 & .1590$\pm$.0029 &
& CRAFT-R & .7593$\pm$.0053 & .4568$\pm$.0079\\
& DyGFormer & .4161$\pm$.0015 & .0811$\pm$.0100 &
& DyGFormer & .7565$\pm$.0079 & .4686$\pm$.0073\\
& DyGFormer+LHA & .6380$\pm$.0071 & .1139$\pm$.0064 &
& DyGFormer+LHA & .6804$\pm$.0085 & .3835$\pm$.0103\\
& TGN & .3969$\pm$.0165 & .0557$\pm$.0106 &
& TGN & .2831$\pm$.0241 & .0933$\pm$.0105\\
& GraphMixer & .4488$\pm$.0042 & .0922$\pm$.0027 &
& GraphMixer & .5128$\pm$.0060 & .2681$\pm$.0059\\
\midrule
Reddit & CRAFT & .8308$\pm$.0117 & .4919$\pm$.0612 &
Wikipedia & CRAFT & .8948$\pm$.0028 & .7366$\pm$.0100\\
& CRAFT-R & .8744$\pm$.0011 & .6637$\pm$.0006 &
& CRAFT-R & .8985$\pm$.0007 & .7101$\pm$.0058\\
& DyGFormer & .8022$\pm$.0029 & .4771$\pm$.0205 &
& DyGFormer & .8283$\pm$.0085 & .5784$\pm$.0094\\
& DyGFormer+LHA & .8647$\pm$.0010 & .6357$\pm$.0070 &
& DyGFormer+LHA & .8522$\pm$.0028 & .6510$\pm$.0102\\
& TGN & .6892$\pm$.0024 & .2204$\pm$.0094 &
& TGN & .6668$\pm$.0248 & .2160$\pm$.0521\\
& GraphMixer & .7088$\pm$.0008 & .2561$\pm$.0009 &
& GraphMixer & .7455$\pm$.0048 & .3160$\pm$.0101\\
\bottomrule
\end{tabular}
\end{table*}

Expected Uniform-20 places CRAFT above DyGFormer on LastFM, whereas the full
catalog places DyGFormer and GraphMixer above CRAFT. On MOOC, the leading
three models are CRAFT, CRAFT-R, and DyGFormer under expected Uniform-20, but
DyGFormer, CRAFT, and CRAFT-R under the full catalog. The Reddit order is
preserved. On Wikipedia, CRAFT and CRAFT-R exchange positions. Most of the
six-model ordering remains stable; the changes concentrate in particular
model pairs.

\subsection{Rank and Module Effect Across \texorpdfstring{\(K\)}{K}}

\begin{table*}[t]
\centering
\caption{Kendall's \(\tau\) between the expected uniform finite-\(K\)
model order and the all-entity order.}
\label{tab:kendall}
\scriptsize
\begin{tabular}{lrrrrr}
\toprule
Dataset & \(K=1\) & \(K=5\) & \(K=20\) & \(K=50\) & \(K=100\)\\
\midrule
LastFM & .733 & .867 & .867 & .867 & 1.000\\
MOOC & .733 & .733 & .733 & 1.000 & 1.000\\
Reddit & 1.000 & 1.000 & 1.000 & 1.000 & 1.000\\
Wikipedia & 1.000 & 1.000 & .867 & .867 & .867\\
\bottomrule
\end{tabular}
\end{table*}

\begin{table*}[t]
\centering
\caption{Effect of CRAFT-R relative to CRAFT, \(G_R(K)\).}
\label{tab:gain}
\scriptsize
\begin{tabular}{lrrrrrr}
\toprule
Dataset & \(K=1\) & \(K=5\) & \(K=20\) & \(K=50\) & \(K=100\) & Full\\
\midrule
LastFM & +.0591 & +.1696 & +.2608 & +.2693 & +.2442 & +.0955\\
MOOC & -.0001 & -.0009 & -.0034 & -.0052 & -.0057 & -.0057\\
Reddit & +.0036 & +.0160 & +.0436 & +.0724 & +.0982 & +.1718\\
Wikipedia & -.0019 & -.0014 & +.0037 & +.0084 & +.0106 & -.0266\\
\bottomrule
\end{tabular}
\end{table*}

In Table~\ref{tab:kendall}, LastFM and MOOC fail to preserve some model-pair
orders at small \(K\). Wikipedia remains at \(\tau=0.867\) for
\(K=20,50,100\). In Table~\ref{tab:gain}, Wikipedia has small positive point
estimates at \(K=20,50,100\), followed by a decrease under the full catalog.
The sign remains positive on Reddit, but the effect size ranges from
\(+0.0034\) at \(K=1\) to \(+0.1718\) over the full catalog. A module
conclusion from sampled ablation is thus conditional on \(q\) and \(K\). In
particular, the Wikipedia result prevents attribution of the sampled point
estimate to a universal effect of repeat-time encoding. Because the module
is fixed and only the candidate protocol changes, the measured gain reflects
an interaction between recurrence encoding and candidate composition.

Table~\ref{tab:mrr_difference_uncertainty} reports uncertainty for the
CRAFT-R minus CRAFT MRR difference computed on matched seeds and test events.
The seed-paired \(t\) interval represents variation across three training
seeds. The time-block paired bootstrap groups chronologically ordered test
events into blocks of 128 and resamples the same blocks jointly for both
models and all seeds. This interval represents test-sequence uncertainty
conditional on trained checkpoints; the seed-paired result is the primary
generalization interval. The Wikipedia Expected-20 gain is \(+0.0038\), with
a seed interval containing zero. The all-entity gain is \(-0.0266\), and
both intervals are below zero. The sampled condition provides a small
positive point estimate with training-seed uncertainty, whereas the
all-entity condition shows a consistent decrease.

\begin{table*}[t]
\centering
\caption{CRAFT-R minus CRAFT MRR differences and 95\% confidence intervals
computed on matched seeds and test events.}
\label{tab:mrr_difference_uncertainty}
\scriptsize
\begin{tabular}{lrrr}
\toprule
Dataset--Protocol & MRR difference & Seed-paired \(t\) CI & Time-block bootstrap CI\\
\midrule
LastFM--Expected-20 & +.2608 & [.2267, .2950] & [.2460, .2756]\\
LastFM--All-entity & +.0955 & [.0845, .1065] & [.0862, .1056]\\
MOOC--Expected-20 & -.0034 & [-.0269, .0200] & [-.0061, -.0010]\\
MOOC--All-entity & -.0057 & [-.0269, .0155] & [-.0107, -.0008]\\
Reddit--Expected-20 & +.0436 & [.0093, .0779] & [.0400, .0471]\\
Reddit--All-entity & +.1718 & [-.0135, .3571] & [.1616, .1813]\\
Wikipedia--Expected-20 & +.0038 & [-.0036, .0112] & [-.0004, .0080]\\
Wikipedia--All-entity & -.0266 & [-.0526, -.0005] & [-.0367, -.0170]\\
\bottomrule
\end{tabular}
\end{table*}

\subsection{Training-Objective Sensitivity}

Table~\ref{tab:objective_sensitivity} reports CRAFT and CRAFT-R retrained
with the BPR objective used by CRAFT. On LastFM and Reddit, the magnitude of
the module effect varies substantially with \(K\) under both BCE and BPR. On
MOOC and Wikipedia, the objective also changes its sign. In particular, the
Expected-20 \(+0.0037\) and all-entity \(-0.0266\) observed for the Wikipedia
BCE checkpoints become \(+0.0345\) and \(+0.0629\), respectively, for the
BPR checkpoints. The sign change under BCE therefore does not generalize as
an objective-independent property of the CRAFT architecture. Within the same
objective and checkpoint, however, both objectives preserve the finding that
\(K\) and the risk set alter the measured module effect.

\begin{table*}[t]
\centering
\caption{CRAFT-R minus CRAFT MRR by training objective. Expected-20 is the
exact expectation of uniform 20-negative evaluation.}
\label{tab:objective_sensitivity}
\small
\begin{tabular}{lrrrr}
\toprule
& \multicolumn{2}{c}{BCE} & \multicolumn{2}{c}{BPR}\\
\cmidrule(lr){2-3}\cmidrule(lr){4-5}
Dataset & Expected-20 & All-entity & Expected-20 & All-entity\\
\midrule
LastFM    & +.2608 & +.0955 & +.2696 & +.0983\\
MOOC      & -.0034 & -.0057 & +.0147 & +.0212\\
Reddit    & +.0436 & +.1718 & +.1035 & +.3431\\
Wikipedia & +.0037 & -.0266 & +.0345 & +.0629\\
\bottomrule
\end{tabular}
\end{table*}

Table~\ref{tab:bpr_membership} applies the same \(2\times2\) diagnostic to the
BPR checkpoints. CRAFT-R obtains .932--1.000 for Repeated--Unseen and
.048--.338 for New--Seen across the four datasets. The large contrast has the
same direction under BCE and BPR, so the mechanistic observation that the
pair-history path conveys membership separation is stable across training
objectives.

\begin{table*}[t]
\centering
\caption{CRAFT-R membership-diagnostic MRR by training objective.
R--U denotes Repeated--Unseen and N--S denotes New--Seen.}
\label{tab:bpr_membership}
\small
\begin{tabular}{lrrrr}
\toprule
& \multicolumn{2}{c}{BCE} & \multicolumn{2}{c}{BPR}\\
\cmidrule(lr){2-3}\cmidrule(lr){4-5}
Dataset & R--U & N--S & R--U & N--S\\
\midrule
LastFM    & .957 & .059 & .978 & .049\\
MOOC      & .888 & .508 & .932 & .338\\
Reddit    & .987 & .048 & 1.000 & .048\\
Wikipedia & .979 & .050 & 1.000 & .048\\
\bottomrule
\end{tabular}
\end{table*}

\begin{figure*}[t]
\centering
\includegraphics[width=0.92\textwidth]{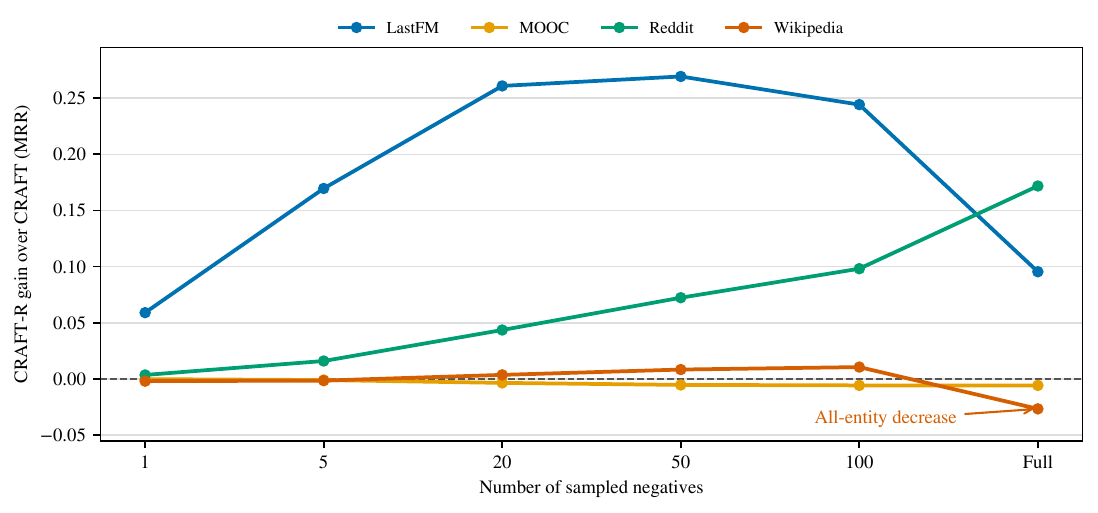}
\caption{CRAFT-R minus CRAFT MRR as a function of the expected number of
negative candidates under uniform sampling. The dashed line marks zero
module effect.}
\label{fig:craft_r_gain_by_k}
\end{figure*}

Figure~\ref{fig:craft_r_gain_by_k} connects the module effects from
Table~\ref{tab:gain} across candidate-set sizes. The horizontal axis gives
the number of uniform random negatives and the full catalog; the vertical
axis gives the MRR difference between CRAFT-R and CRAFT, with values above
the dashed line favoring CRAFT-R. Effect size changes substantially with
\(K\) on LastFM and Reddit. On Wikipedia, small positive point estimates at
sampled \(K=20,50,100\) move to a negative full-catalog effect. The curve
visualizes how the measured effect of an unchanged module varies
continuously with candidate count and why a sampled ablation conclusion need
not represent the full-catalog conclusion.

\subsection{Representation Controls and Risk-Set Sensitivity}

\begin{table*}[t]
\centering
\caption{Full-catalog MRR after removing representation directions from
CRAFT-R.}
\label{tab:projection}
\scriptsize
\begin{tabular}{lrrrr}
\toprule
Dataset & Base & Membership & Random & Shuffled\\
\midrule
LastFM & .1590 & .0738 & .1588 & .1570\\
MOOC & .4568 & .3278 & .4600 & .4605\\
Reddit & .6637 & .0416 & .6640 & .6537\\
Wikipedia & .7101 & .0405 & .7206 & .7121\\
\bottomrule
\end{tabular}
\end{table*}

In Table~\ref{tab:projection}, removing a random direction or a
shuffled-label direction preserves the baseline result. Removing the
membership direction produces consistently different outcomes on all four
datasets. The controls distinguish removal of an arbitrary representation
dimension from intervention on a representation axis carrying pair
membership.

Model order is identical under the fixed training catalog and the
time-available catalog. Their MRR differences are zero on LastFM and MOOC,
0.0002--0.0005 on Reddit, and 0.0020--0.0054 on Wikipedia.

\subsection{Evaluation Cost}

\begin{figure*}[t]
\centering
\includegraphics[width=0.92\textwidth]{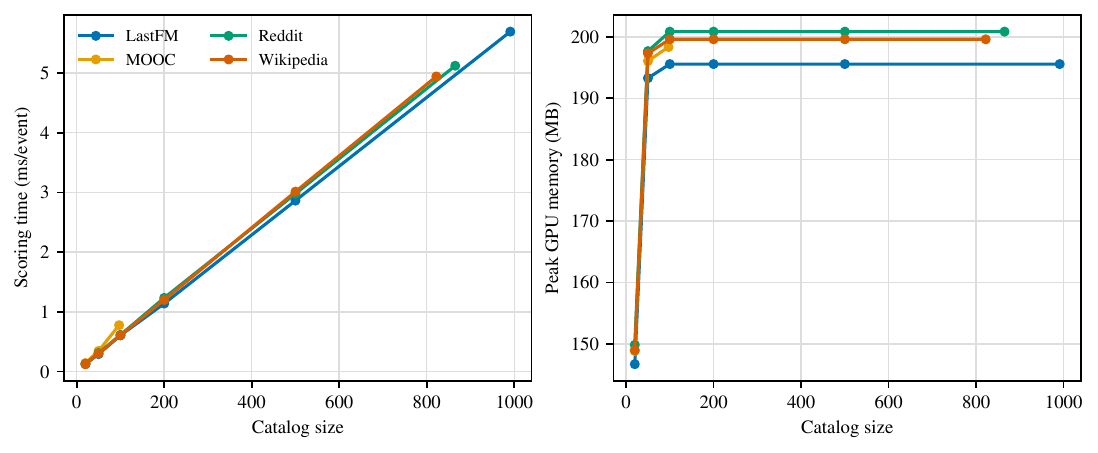}
\caption{All-entity scoring time and peak GPU memory versus fixed-catalog
size. We evaluate 128 test events from each CRAFT checkpoint in candidate
batches of 4096; points show the mean over three seeds.}
\label{fig:catalog_scaling}
\end{figure*}

Figure~\ref{fig:catalog_scaling} reports all-entity scoring cost as catalog
size grows for the same CRAFT checkpoint. With 128 test events and candidate
batches of 4096, time per event increases nearly linearly from
0.13--0.14\,ms for approximately 20 candidates to 4.94--5.69\,ms for
822--991 candidates. Candidate batching makes peak GPU memory saturate near
200\,MB. Exact ranking is therefore directly computable at the catalog scale
studied here; this result does not represent the cost of direct evaluation
for million-entity catalogs.

\section{Discussion and Conclusion}
\label{sec:conclusion}

A sampled-negative CTDG score is jointly produced by the model scorer and the
candidate protocol. A non-uniform \(q\) changes the Bayes-optimal ordering
through a density ratio. Uniform \(q\) over a fixed catalog preserves the
population-optimal order, but finite \(K\) observes only part of the complete
ranking and does not reliably preserve model order or module gain. This
distinction decomposes the familiar observation that ``random negatives are
easy'' into two precise evaluation effects.

Pair recurrence is useful information in the actual data-generating process.
When positive events concentrate on seen pairs relative to their prevalence
in a uniform catalog, the optimal membership coefficient is positive. The
evaluation issue arises when this coarse subgroup separation accounts for a
large share of MRR in a small candidate set and obscures destination
selection within the seen and unseen groups. A support-restricted sampler
such as source-history can create perfect separation in the opposite
direction. The \(2\times2\) evaluation and representation intervention
separate subgroup separation, recurrent-link ranking, and novel-link ranking,
and show how pair-history operations transmit this difference to the score.

This distinction supports a mechanistic attribution of CRAFT's performance.
CRAFT's candidate-to-history cross-attention computes semantic compatibility
and a match with historical pair identity along the same path. Under random
candidates, both signals frequently identify the answer in the same
direction, so high sampled MRR measures their combined effect. Crossing
membership composition and removing the corresponding representation axis
show that pair membership accounts for the large amplitude of the subgroup
diagnostics. Membership-matched and all-entity rankings provide controlled
conditions measuring residual ranking performance after membership-aligned
separation.

The lower absolute MRR of the full catalog relative to expected Uniform-20
is an arithmetic consequence of having more candidates. Changes in
comparative conclusions occur for particular model pairs and modules. At
least one model pair changes relative order on LastFM, MOOC, and Wikipedia.
The CRAFT-R module effect on Wikipedia has a small positive sampled point
estimate with seed-level uncertainty, but a consistently negative
full-catalog value under the unified BCE objective. With BPR retraining, both
sampled and all-entity effects are positive. The training objective changes
the scores supplied by an architecture, while the candidate protocol changes
the comparison problem used to measure those scores. Architecture
superiority and ablation effects supported by sampled-negative results are
therefore reproducible conditional evidence when \(q\) and \(K\) are stated
together. The measured effect of repeat-time encoding is accordingly an
interaction between the module and candidate protocol rather than a fixed
property of the module alone.

Fixed-training-catalog all-entity ranking applies directly to any existing
CTDG model that provides source--destination scores. It eliminates freedom
in selecting negatives and removes sampling variance, measuring one CTDG
destination-prediction problem specified by the catalog definition and its
coverage. Historical, inductive, and source-history candidates remain useful
diagnostics for dissecting model behavior. Placing all-entity results as the
primary architecture and module comparison, with sampled protocols reported
as secondary analyses that explicitly state \(q\), \(K\), and support,
stabilizes interpretation. This recommendation applies to benchmarks with
an exactly enumerable fixed catalog. At large scale, reporting retrieval
recall and approximate-ranking error against the same fixed all-entity
target preserves the evaluation question.

All-entity ranking fixes the evaluation target by removing candidate
selection at test time. The negative sampler and loss function used during
training remain part of the learned model specification. The BPR sensitivity
results demonstrate the need to report the training objective together with
the evaluation protocol.

\section*{Acknowledgements}
The authors have no acknowledgements to declare.

\section*{Funding}
This research received no specific grant from funding agencies in the public,
commercial, or not-for-profit sectors.

\section*{Data availability}
This study uses the public Wikipedia, Reddit, MOOC, and LastFM temporal
interaction datasets distributed with JODIE. The raw data are available from
the Stanford SNAP JODIE repository. The accompanying materialization pipeline
records the source metadata and produces the chronological splits used in this
study.

\section*{Code availability}
The code for data materialization, training, sampled-candidate diagnostics,
\(2\times2\) controls, fixed-catalog all-entity evaluation, and result
reproduction is publicly available at
\url{https://github.com/SnowyPainter/goodctdg-public}.

\bibliographystyle{IEEEtran}
\bibliography{references}

\appendix
\label{app:reproducibility}

\textbf{Additional Reproducibility Information.}

Every sampled-\(K\) and full-catalog result is derived from one score vector
computed per event. Training stochasticity is independent across seeds.
Within a seed, the candidates for different \(K\) form nested prefixes of one
fixed random permutation. The complete experiment comprises 72 training and
evaluation runs: four datasets, six models, and three seeds.

For paired-random and source-history comparisons, both conditions use only
events for which 20 source-history negatives can be constructed. The
\(2\times2\) evaluation uses the same positive events for which all four
conditions are available. The numbers of repeated and new events are 14,979
and 3,387 on LastFM; 698 and 273 on Wikipedia; 729 and 151 on Reddit; and
6,106 and 3,981 on MOOC.

\subsection{Derivation of the Uniform Finite-\texorpdfstring{\(K\)}{K}
Expectation}

If \(r-1\) catalog candidates score above the positive, the probability mass
function in Eq.~\eqref{eq:hypergeom_rank} gives
\begin{equation}
\mathbb E[\operatorname{RR}_K\mid r,N]
=\frac{1}{\binom{N-1}{K}}
\sum_x\frac{\binom{r-1}{x}\binom{N-r}{K-x}}{x+1}.
\label{eq:expected_rr_sum}
\end{equation}
Applying
\(\binom{r-1}{x}/(x+1)=\binom{r}{x+1}/r\) and simplifying the sum with
Vandermonde's identity yields
\(\{\binom{N}{K+1}-\binom{N-r}{K+1}\}/
\{r\binom{N-1}{K}\}\), giving
Eq.~\eqref{eq:expected_sampled_rr}.

We use average rank for ties. Let \(a\) and \(b\) be the numbers of candidates
scoring strictly above and equal to the positive, and let \(H\) and \(T\)
denote the corresponding numbers included in a sample. The tie-aware
reciprocal rank is \(1/(1+H+T/2)\), implemented as
\begin{equation}
\sum_{h,t}
\frac{\binom{a}{h}\binom{b}{t}
\binom{N-1-a-b}{K-h-t}}{\binom{N-1}{K}}
\,\frac{1}{1+h+t/2}.
\label{eq:tie_aware_expectation}
\end{equation}
Only three of 352,458 model--event instances across the 72 BCE runs contain
ties (0.00085\%).

\end{document}